\documentclass{article}
\usepackage[utf8]{inputenc}

\title{\href{https://www.kaggle.com/mi2datalab/mementoml}{MementoML}: Performance of selected machine learning algorithm configurations on OpenML100 datasets}

\date{July 2020}

\usepackage{natbib}
\setcitestyle{authoryear,open={[},close={]}}
\usepackage{authblk}
\usepackage{graphicx}
\usepackage{hyperref}
\usepackage{booktabs}
\usepackage{multirow}
\usepackage[table,xcdraw]{xcolor}

\usepackage[margin=1.1in]{geometry}

\usepackage{float}
\restylefloat{table}
\author[1]{Wojciech Kretowicz}
\author[1,2]{Przemysław Biecek}

\affil[1]{Warsaw University of Technology}
\affil[2]{University of Warsaw}

\begin{document}

\maketitle

\section{Introduction}
Finding optimal hyperparameters for the machine learning algorithm can often significantly improve its performance \cite{probst2018tunability}. But how to choose them in a time-efficient way? In this paper we present the protocol of generating benchmark data describing the performance of different ML algorithms with different hyperparameter configurations. Data collected in this way is used to study the factors influencing the algorithm's performance.

This collection was prepared for the purposes of the study presented in the \textbf{EPP} paper \cite{gosiewska2020interpretable}. 
We tested algorithms performance on dense grid of hyperparameters. Tested datasets and hyperparameters were chosen before any algorithm has run and were not changed.
This is a different approach than the one usually used in hyperparameter tuning, where the selection of candidate hyperparameters depends on the results obtained previously.
However, such selection allows for systematic analysis of performance sensitivity from individual hyperparameters.

This resulted in a comprehensive dataset of such benchmarks that we would like to share. We hope, that computed and collected result may be helpful for other researchers. This paper describes the way data was collected. Here you can find benchmarks of 7 popular machine learning algorithms on 39 OpenML datasets.

The detailed data forming this benchmark are available at: \url{https://www.kaggle.com/mi2datalab/mementoml}.

\section{Related datasets}

\cite{khn2018automatic} introduced  a benchmark of algorithms created for the OpenML repository. This dataset contains data about 6 algorithms written in \textbf{R}: glmnet, rpart, kknn, svm, ranger, xgboost. It also allows us to run some additional computations and obtain further results in a similar way.

The \cite{smith2014easy} provides a MongoDB database that has got data on an instance level. It contains predictions made for every single instance in the considered datasets, information about algorithms and its hyperparameters. There is also a possibility to extend this benchmark. 

\textit{mlpack} benchmark \cite{edel2014automatic} contains data about performance of different algorithms in popular machine learning frameworks/libraries. It also provides comprehensive scripts for further evaluation.

\section{Algorithms, datasets and hyperparameters used}

We used several number of popular machine learning algorithms: \textbf{gradient boosting on decision trees} (\textit{catboost} \cite{prokhorenkova2017catboost}, \textit{gbm}, \textit{xgboost}), \textbf{generalized linear models} (\textit{glmnet} \cite{glmnet}), \textbf{$k$ nearest neighbours} (\textit{kknn}), \textbf{random forests} (\textit{randomforest} \cite{randomforest}, \textit{ranger} \cite{ranger}). All computations were made in \textbf{R} and all models were from \textbf{R} packages. Almost all of them are used through the \textbf{mlr} \cite{mlr} framework. Only \textit{catboost} was used directly, because is not included in \textbf{mlr}.

Benchmarks on predicted values were calculated with \textbf{mlr} function \textbf{performance} for all models except \textbf{catboost}. For that one were used \textbf{measureACC} and \textbf{measureAUC}.

\subsection{Algorithms and parameters}

These are models used in the computation with considered ranges of parameters:

\begin{table}[H]
\centering
\caption{Algorithms, their parameters' ranges used in benchmark and general rule of drawing. $U$ stands for random variable sampled from uniform distribution on proper set.}
\label{params}
\begin{tabular}{@{}llllll@{}}
\toprule
\rowcolor[HTML]{656565} 
{\color[HTML]{FFFFFF} algorithm}                       & {\color[HTML]{FFFFFF} parameter} & {\color[HTML]{FFFFFF} type} & {\color[HTML]{FFFFFF} lower} & {\color[HTML]{FFFFFF} upper} & {\color[HTML]{FFFFFF} grid} \\ \midrule
\cellcolor[HTML]{C0C0C0}                               & iterations                       & integer                     & 100                          & 10086                        & $2^U$                       \\
\cellcolor[HTML]{C0C0C0}                               & depth                            & integer                     & 6                            & 10                           & $U$                         \\
\cellcolor[HTML]{C0C0C0}                               & l2\_leaf\_reg                    & numeric                     & 0                            & 9                            & $U^2$                         \\
\cellcolor[HTML]{C0C0C0}                               & bagging\_temperature             & numeric                     & 0                            & 1.5e                         & $U$                         \\
\multirow{-5}{*}{\cellcolor[HTML]{C0C0C0}catboost}     & learning\_rate                   & numeric                     & 0.001                        & 2                            & $2^U$                       \\ \midrule
\cellcolor[HTML]{C0C0C0}                               & n.trees                          & integer                     & 100                          & 10086                        & $2^U$                       \\
\cellcolor[HTML]{C0C0C0}                               & interaction.depth                & integer                     & 1                            & 5                            & $U$                         \\
\cellcolor[HTML]{C0C0C0}                               & n.minobsinnode                   & integer                     & 2                            & 25                           & $U$                         \\
\cellcolor[HTML]{C0C0C0}                               & shrinkage                        & numeric                     & 0.001                        & 0.1                          & $10^U$                      \\
\multirow{-5}{*}{\cellcolor[HTML]{C0C0C0}gbm}          & bag.fraction                     & numeric                     & 0.2                          & 1                            & $U$                         \\ \midrule
\cellcolor[HTML]{C0C0C0}                               & alpha                            & numeric                     & 0                            & 1                            & $U$                         \\
\multirow{-2}{*}{\cellcolor[HTML]{C0C0C0}glmnet}       & lambda                           & numeric                     & 0.001                        & 1024                         & $2^U$                       \\ \midrule
\cellcolor[HTML]{C0C0C0}kknn                           & k                                & integer                     & 1                            & 30                           & $U$                         \\ \midrule
\cellcolor[HTML]{C0C0C0}                               & ntree                            & integer                     & 100                          & 10086                        & $2^U$                       \\
\cellcolor[HTML]{C0C0C0}                               & replace                          & logical                     & FALSE                        & TRUE                         & $U$                         \\
\multirow{-3}{*}{\cellcolor[HTML]{C0C0C0}randomforest} & nodesize                         & integer                     & 1                            & 5                            & $U$                         \\ \midrule
\cellcolor[HTML]{C0C0C0}                               & num.trees                        & integer                     & 100                          & 10086                        & $2^U$                       \\
\cellcolor[HTML]{C0C0C0}                               & min.node.size                    & integer                     & 1                            & 4                            & $U$                         \\
\cellcolor[HTML]{C0C0C0}                               & replace                          & logical                     & FALSE                        & TRUE                         & $U$                         \\
\multirow{-4}{*}{\cellcolor[HTML]{C0C0C0}ranger}       & splitrule                        & discrete                    & -                            & -                            & $U$                         \\ \midrule
\cellcolor[HTML]{C0C0C0}                               & booster                          & discrete                    & -                            & -                            & $U$                         \\
\cellcolor[HTML]{C0C0C0}                               & nrounds                          & integer                     & 1                            & 1000                         & $U$                         \\
\cellcolor[HTML]{C0C0C0}                               & eta                              & numeric                     & 0.031                        & 1                            & $2^U$                       \\
\cellcolor[HTML]{C0C0C0}                               & subsample                        & numeric                     & 0.5                          & 1                            & $U$                         \\
\cellcolor[HTML]{C0C0C0}                               & max\_depth                       & integer                     & 6                            & 15                           & $U$                         \\
\cellcolor[HTML]{C0C0C0}                               & min\_child\_weight               & numeric                     & 1                            & 8                            & $2^U$                       \\
\cellcolor[HTML]{C0C0C0}                               & colsample\_bytree                & numeric                     & 0.2                          & 1                            & $U$                         \\
\multirow{-8}{*}{\cellcolor[HTML]{C0C0C0}xgboost}      & colsample\_bylevel               & numeric                     & 0.2                          & 1                            & $U$                         \\ \bottomrule
\end{tabular}
\end{table}

Parameter \textbf{splitrule} for \textbf{ranger} was either \textbf{gini} or \textbf{extratrees}. Parameter \textbf{booster} for \textbf{xgboost} was either \textbf{gbtree} or \textbf{gblinear}.

Parameters for each model were randomly chosen within the presented ranges using corresponding distribution. Although they were drawn randomly, all of them are reproducible.

\subsection{Datasets}

All datasets used in the benchmark were downloaded from \textbf{OpenML} \cite{OpenML2013}. These datasets come from \textbf{OpenML100} and are all for binary classification, the number of observations are between $500$ and $100 000$, the number of features is less than $5000$, and the ratio of the minority class and the majority class is above $0.05$.

All considered datasets' \textbf{OpenML} ids:

\begin{table}[H]
\begin{tabular}{
>{\columncolor[HTML]{C0C0C0}}l llll}
\cellcolor[HTML]{656565}{\color[HTML]{FFFFFF} id} & \cellcolor[HTML]{656565}{\color[HTML]{FFFFFF} name} & \cellcolor[HTML]{656565}{\color[HTML]{FFFFFF} link} & \cellcolor[HTML]{656565}{\color[HTML]{FFFFFF} rows} & \cellcolor[HTML]{656565}{\color[HTML]{FFFFFF} columns} \\
3                                                 & kr-vs-kp                                            & \url{https://www.openml.org/d/3}                          & 3196                                                & 37                                                     \\
31                                                & credit-g                                            & \url{https://www.openml.org/d/31}                         & 1000                                                & 21                                                     \\
37                                                & diabetes                                            & \url{https://www.openml.org/d/37}                         & 768                                                 & 9                                                      \\
44                                                & spambase                                            & \url{https://www.openml.org/d/44}                         & 4601                                                & 58                                                     \\
50                                                & tic-tac-toe                                         & \url{https://www.openml.org/d/50}                         & 958                                                 & 10                                                     \\
151                                               & electricity                                         & \url{https://www.openml.org/d/151}                        & 45312                                               & 9                                                      \\
312                                               & scene                                               & \url{https://www.openml.org/d/312}                        & 2407                                                & 300                                                    \\
333                                               & monks-problems-1                                    & \url{https://www.openml.org/d/333}                        & 556                                                 & 7                                                      \\
334                                               & monks-problems-2                                    & \url{https://www.openml.org/d/334}                        & 601                                                 & 7                                                      \\
335                                               & monks-problems-3                                    & \url{https://www.openml.org/d/335}                        & 554                                                 & 7                                                      \\
1036                                              & sylva\_agnostic                                     & \url{https://www.openml.org/d/1036}                       & 14395                                               & 217                                                    \\
1038                                              & gina\_agnostic                                      & \url{https://www.openml.org/d/1038}                       & 3468                                                & 971                                                    \\
1043                                              & ada\_agnostic                                       & \url{https://www.openml.org/d/1043}                       & 4562                                                & 49                                                     \\
1046                                              & mozilla4                                            & \url{https://www.openml.org/d/1046}                       & 15545                                               & 6                                                      \\
1049                                              & pc4                                                 & \url{https://www.openml.org/d/1049}                       & 1458                                                & 38                                                     \\
1050                                              & pc3                                                 & \url{https://www.openml.org/d/1050}                       & 1563                                                & 38                                                     \\
1063                                              & kc2                                                 & \url{https://www.openml.org/d/1063}                       & 522                                                 & 22                                                     \\
1067                                              & kc1                                                 & \url{https://www.openml.org/d/1067}                       & 2109                                                & 22                                                     \\
1068                                              & pc1                                                 & \url{https://www.openml.org/d/1068}                       & 1109                                                & 22                                                     \\
1120                                              & MagicTelescope                                      & \url{https://www.openml.org/d/1120}                       & 19020                                               & 12                                                     \\
1461                                              & bank-marketing                                      & \url{https://www.openml.org/d/1461}                       & 45211                                               & 17                                                     \\
1462                                              & banknote-authentication                             & \url{https://www.openml.org/d/1462}                       & 1372                                                & 5                                                      \\
1464                                              & blood-transfusion-service-center                    & \url{https://www.openml.org/d/1464}                       & 748                                                 & 5                                                      \\
1467                                              & climate-model-simulation-crashes                    & \url{https://www.openml.org/d/1467}                       & 540                                                 & 21                                                     \\
1471                                              & eeg-eye-state                                       & \url{https://www.openml.org/d/1471}                       & 14980                                               & 15                                                     \\
1479                                              & hill-valley                                         & \url{https://www.openml.org/d/1479}                       & 1212                                                & 101                                                    \\
1480                                              & ilpd                                                & \url{https://www.openml.org/d/1480}                       & 583                                                 & 11                                                     \\
1485                                              & madelon                                             & \url{https://www.openml.org/d/1485}                       & 2600                                                & 501                                                    \\
1486                                              & nomao                                               & \url{https://www.openml.org/d/1486}                       & 34465                                               & 119                                                    \\
1487                                              & ozone-level-8hr                                     & \url{https://www.openml.org/d/1487}                       & 2534                                                & 73                                                     \\
1489                                              & phoneme                                             & \url{https://www.openml.org/d/1489}                       & 5404                                                & 6                                                      \\
1494                                              & qsar-biodeg                                         & \url{https://www.openml.org/d/1494}                       & 1055                                                & 42                                                     \\
1504                                              & steel-plates-fault                                  & \url{https://www.openml.org/d/1504}                       & 1941                                                & 34                                                     \\
1510                                              & wdbc                                                & \url{https://www.openml.org/d/1510}                       & 569                                                 & 31                                                     \\
1570                                              & wilt                                                & \url{https://www.openml.org/d/1570}                       & 4839                                                & 6                                                      \\
4134                                              & Bioresponse                                         & \url{https://www.openml.org/d/4134}                       & 3751                                                & 1777                                                   \\
4135                                              & Amazon\_employee\_access                            & \url{https://www.openml.org/d/4135}                       & 32769                                               & 10                                                     \\
4534                                              & PhishingWebsites                                    & \url{https://www.openml.org/d/4534}                       & 11055                                               & 31                                                     \\
40509                                             & Australian                                          & \url{https://www.openml.org/d/40509}                      & 690                                                 & 15                                                    
\end{tabular}
\end{table}

\section{Data collection}

Each dataset was divided into $20$ \textit{train}/\textit{test} bootstrap pairs. Each row is not guaranteed to be chosen exactly once to the test set or $19$ times to the train set because each pair was chosen independently in a bootstrap manner. Each considered model with a particular set of hyperparameters was $20$ times trained, one time on each \textit{train} subset and evaluated on a corresponding \textit{test}. There were two collated measures: \textbf{ACC} and \textbf{AUC}. Thus, for each model and each dataset should be $20\cdot |paramset|$ \textbf{ACC} and $20\cdot |paramset|$ \textbf{AUC} measures. However, not all computations finished and they are updated on a regular basis as they progress. Additionally, learning times were collected.

What is important in our approach, the first thing we did is arbitrary choosing datasets and randomly choosing parameters in a way described above in a table \ref{params}. Thus they had been all chosen and set before any calculations began and they were not updated meanwhile. Splits to \textit{train/test} subsets are also constant for each dataset. This makes comparing results between different algorithms easier. We are trying to cover as much of parameter space as possible, including also parameters in subspaces that may not work well or even work at all. This should make possible researches in more broad-spectrum concerning hyperparameters. However, for practical reasons, we had to draw them only from finite ranges.

\subsection{ABC4ML - Automated benchmark collector}

\textbf{ABC4ML} is an abbreviation of \textbf{Automated benchmark collector}, software written and used for easy and reproducible benchmarking of selected datasets. Its main function is \textit{calculate} that simply takes model name, vector of \textbf{OpenML} datasets' ids, path to a file with \textit{train}/\textit{test} splits and a path to file with parameters sets. This allows easy and transparent calculations as well as easy parallelization. If considered machine learning algorithm did not converge there would be returned \textit{NA}.

As you see, you need to create a file with data splits in a form of rows ids and file with parameters sets first, before any calculations.

Due to use of OpenML, there is no need for downloading these datasets by hand prior to run calculations. They are one by one downloaded to RAM as they are needed in progressing computations.

\subsection{Details}

After some initial data collection, we obtained estimated times of calculations on each dataset. Then datasets were grouped in such a way, that sum of time required for all datasets in a group is nearly equal with other groups.

For each model and each group of datasets was run a docker container that was calculating benchmarks.

\subsection{Dataset}

\subsubsection{Benchmarks}

Resulting dataset is a dataframe with 7 columns. First "dataset" column denotes \textbf{OpenML} id of the dataset i.e. \textit{1486}, next "row\_index" is the train/test split identifier i.e. \textit{12} from splits file, third "model" is a model name i.e. \textit{gbm} or \textit{kknn}. Fourth is a "param\_index" denoting hyperparameter set identifier from parameters file. These identifiers starts from 1001 (1001 denotes 1). Fifth is "time", that is a learning time measured in \textit{ms}. The last two columns are \textit{acc} and \textit{auc} measures.

\subsubsection{Hyperparameters}

There is also a data frame for each model with its hyperparameters. The first column is a "param\_index". Rest of the columns correspond to hyperparameters related to this model and used in calculations.

\subsubsection{Train/test splits}

For each used dataset there is a separate file with a train/test splits. Each of its rows indicates row indices in a single test subset in the mentioned dataset.

\section{Reproducibility}

All hyperparameters were chosen with a fixed random seed, thus they can be reproduced at will. However, not all parameters were chosen in the same way. \textbf{catboost} parameters were drawn using a newer version of the script. In this updated version of the scipt, there is also a possibility to draw parameters for other algorithms.

Similarly, all algorithms had fixed random seeds before every single run (default seed parameter in a \textit{calculate} function). Thus you can easily and independently reproduce only some of the results without the need to make all previous calculations. Some of the results were reproduced to ensure the proper functioning of the whole framework.

Moreover, we trace versions of the used software libraries.

\section{Further calculations and Docker}

It is much easier to make calculations by your own using our docker container. This container is a modified \textbf{r-base} image that installs all needed both \textbf{linux} and \textbf{R} packages, has a structure compliance with presumptions of a \textit{calculate.R} script, has got a \textit{screen} utility, copies train/test splits from "datasets" directory and parameters from "parameters". If you want to add some new algorithm that is not present in \textbf{mlr} framework, you need to add an install statement in an \textit{install\_packages.R} script and add new function to \textit{calculate.R} with compliant result vector.

To build docker just run:

\texttt{sudo docker build .}

Before you run docker ensure you have a directory for results in a host. \textit{results} directory inside a container has to be matched with the host's results directory.

To run a docker simply type:

\texttt{sudo docker run -ti -v [host's results absolute directory]:/results elo bash}

Open \textbf{R} console after running this command, source \textit{calculate.R} function from "scripts" and run it with proper parameters.

All results will be saved to host's result directory passed in a \texttt{docker run} command.

\section{Discussion}
This work results in a comprehensive dataset with wide hyperparameter space covered. Dataset is ready to use, easily accessible and, if needed, allows further calculations in compliance form. It can be used in developing a branch of machine learning - meta-learning, finding the best defaults of hyperparameters and reasonable subspaces as well as discovering the impact and importance of particular ones.

This dataset can be found at: \href{https://www.kaggle.com/mi2datalab/mementoml}{MementoML}

\bibliography{references}
\end{document}